\documentclass[journal]{IEEEtran}

\ifCLASSINFOpdf
\else
   \usepackage[dvips]{graphicx}
\fi
\usepackage{amsmath}
\usepackage[hyphens]{url}
\usepackage{cite}
\usepackage{balance}
\usepackage{ulem}
\usepackage{multirow}
\usepackage{graphicx}
\usepackage{booktabs}
\usepackage[acronym]{glossaries}
\usepackage[caption=false,farskip=0pt]{subfig}
\newacronym{bisst}{Bi-SST}{bidirectional single-stream}
\newacronym{bert}{BERT}{Bidirectional Encoder Representations from Transformers}
\newacronym{bmt}{BMT}{Bi-Modal Transformer}
\newacronym{c3d}{C3D}{Convolutional 3D Network}
\newacronym{cnn}{CNN}{Convolutional Neural Network}
\newacronym{dap}{DAP}{Deep Action Proposal}
\newacronym{ed}{ED}{Elaborative Descriptions}
\newacronym{fps}{FPS}{frames per second}
\newacronym{gcn}{GCN}{Graph Convolutional Network}
\newacronym{glove}{GloVE}{Global Vectors}
\newacronym{gan}{GAN}{Generative Adversarial Network}
\newacronym{gru}{GRU}{Gated Recurrent Unit}
\newacronym{har}{HAR}{Human Action Recognition}
\newacronym{i3d}{I3D}{Inflated 3D Network}
\newacronym{idt}{IDT}{Improved Dense Trajectories}
\newacronym{lstm}{LSTM}{Long Short-Term Memory}
\newacronym{mdvc}{MDVC}{Multi-modal Dense Video Captioning}
\newacronym{mse}{MSE}{Mean Squared Error}
\newacronym{nlp}{NLP}{Natural Language Processing}
\newacronym{of}{OF}{Optical Flow}
\newacronym{rnn}{RNN}{Recurrent Neural Network}
\newacronym{sbert}{SBERT}{Sentence-BERT}
\newacronym{sota}{SOTA}{state-of-the-art}
\newacronym{svm}{SVM}{Support Vector Machine}
\newacronym{tac}{TAC}{trimmed action classification}
\newacronym{zsar}{ZSAR}{Zero-shot Action Recognition}
\newacronym{zsl}{ZSL}{Zero-shot Learning}

\DeclareMathOperator*{\argmax}{arg\,max}
\usepackage[dvipsnames,table,xcdraw]{xcolor}

\newacronymstyle{long-short-br}
{%
  \GlsUseAcrEntryDispStyle{long-short}%
}%
{%
  \GlsUseAcrStyleDefs{long-short}%
}
\setacronymstyle{long-short-br}

\newif\ifalt
\alttrue

\ifalt
\usepackage[colorlinks=true,allcolors=black]{hyperref}
\else
\fi

\usepackage[capitalise]{cleveref} %

\begin{document}

\title{Global Semantic Descriptors for\\Zero-Shot Action Recognition}

\author{Valter Estevam, Rayson Laroca, Helio Pedrini, and David Menotti
\thanks{This work was supported by the Federal Institute of Paran\'{a}, Federal University of Paran\'{a} and by grants from the National Council for Scientific and Technological Development (CNPq)
(grant numbers 309330/2018-1 and 308879/2020-1). The Titan Xp and Quadro RTX $8000$ GPUs used for this research were donated by the NVIDIA Corporation.}
\thanks{Valter Estevam is with the Federal Institute of Paran\'a, Irati 84500-000, Brazil, and also with the Federal University of Paraná, Curitiba 81531-970, Brazil (e-mail: \textit{valter.junior@ifpr.edu.br})}
\thanks{Rayson Laroca and David Menotti are with the Federal University of Paran\'a, Curitiba 81531-970, Brazil (e-mail: \textit{rblsantos@inf.ufpr.br}, \textit{menotti@inf.ufpr.br})}
\thanks{Helio Pedrini is with the University of Campinas, Campinas 13083-852, Brazil (e-mail: \textit{helio@ic.unicamp.br})}
\ifalt
\thanks{This is an author-prepared version. The published version is available at the \textit{IEEE Xplore Digital Library} (DOI: \href{http://doi.org/10.1109/LSP.2022.3200605}{\textcolor{blue}{10.1109/LSP.2022.3200605}}).}
\else
\fi
}
\maketitle

\begin{abstract}
The success of \gls*{zsar} methods is intrinsically related to the nature of semantic side information used to transfer knowledge, although this aspect has not been primarily investigated in the literature. This work introduces a new ZSAR method based on the relationships of actions-objects and actions-descriptive sentences. We demonstrate that representing all object classes using descriptive sentences generates an accurate object-action affinity estimation when a paraphrase estimation method is used as an embedder. We also show how to estimate probabilities over the set of action classes based only on a set of sentences without hard human labeling. In our method, the probabilities from these two global classifiers (\textit{i.e.}, which use features computed over the entire video) are combined, producing an efficient transfer knowledge model for action classification. Our results are state-of-the-art in the Kinetics-400 dataset and are competitive on UCF-101 under the \gls*{zsar} evaluation.
\ifalt
Our code is available at \textit{\url{https://github.com/valterlej/objsentzsar}}. 
\else
Our code is available at \url{https://github.com/valterlej/objsentzsar}. 
\fi
\end{abstract}


\IEEEpeerreviewmaketitle

\section{Introduction}

\glsresetall

\IEEEPARstart{D}{eep} learning has been applied in \gls*{har} in videos with remarkable results in the last decade~\cite{carreira:2017,basak:2022}.
Deep models require many annotated samples for each class we want to classify, typically hundreds of videos.
Currently, Kinetics-700~\cite{carreira:2019} is the largest \gls*{har} dataset, with 700 action classes and at least 700 videos per class, totaling 647,907.
Even considering this large number of actions, numerous more are to be collected and annotated in the real world, demanding intensive human labor and retraining supervised models with the new data.
These limitations in the supervised learning paradigm motivate the \gls*{zsar}~problem.

A \gls*{zsar} method aims to  classify samples from unknown classes, \textit{i.e.,} classes that were unavailable in the model training phase. This goal can only be achieved by transferring knowledge from other models and adding semantic information~\cite{estevam:2021survey}. 
Usually, the videos are embedded by off-the-shelf \glspl*{cnn} (\textit{e.g.},
C3D~\cite{tran:2015}, i3D~\cite{carreira:2017}), and the labels are encoded by attributes or word vectors (\textit{e.g.}, Word2Vec~\cite{mikolov:2013}, GloVe~\cite{pennington:2014} or Fast Text~\cite{grave:2018}).
As shown in~\cite{estevam:2021survey}, methods based on attributes frequently perform better than versions based on deep encoding. 
Nevertheless, annotating classes with attributes is not scalable.
A strategy to overcome the limitation imposed by human annotation is to take a set of objects as attributes and pre-compute descriptors in a semantic space~\cite{jain:2015,mettes:2021,bretti:2021,mettes:2022}.
Hence, we can recognize a set of objects in a video (\textit{e.g.}, using a pre-trained~\gls*{cnn}) and infer the most compatible human~action.

For example, Jain~\textit{et al.}~\cite{jain:2015} introduced a method to relate objects and actions by incorporating semantic information in the form of object labels encoded with Word2Vec embeddings improved by Gaussian mixtures.
In their approach, a set of objects is recognized by selecting frames from the videos and averaging the object probability estimations from a \gls*{cnn} pre-trained on ImageNet~\cite{krizhevsky:2017}. Posteriorly, Mettes and Snoek~\cite{mettes:2017} introduced the concept of spatial-aware object embeddings in which an action signature is computed by locating objects and humans.
Their label encoding was computed with~Word2Vec.

Bretti and Mettes~\cite{bretti:2021}, on the other hand, proposed a method to improve the predictions of objects by considering object-scene compositions.
They also employed~\gls*{sbert}~(as used in~\cite{chen:2021}~and~\cite{estevam:2021tell}) to compute sentence embeddings over object-scene label compositions.
However, unlike us, they did not observe a significant improvement compared to adopting word embeddings (using Fast Text), probably because they did not provide sufficient semantic information to the model.
Finally, Mettes~\textit{et al.}~\cite{mettes:2021} investigated some prior knowledge such as person/object location and spatial relation, expanding previous works~\cite{mettes:2017,bretti:2021}.
They also investigated semantic ambiguity by adopting label embeddings in languages other than English.

Estevam~\textit{et al.}~\cite{estevam:2021tell} demonstrated that the automatic generation of sentences employing video captioning models~\cite{estevam2021dense} can be used as a significant global semantic descriptor providing information on actors, objects, scenes, and their relationships.
They also demonstrated how important it is to represent actions not with a single label (\textit{e.g.,}~\cite{jain:2015,mettes:2017}), nor with a single or a few sentences (\textit{e.g.,}~\cite{chen:2021}), but with one or two dozens of descriptive sentences 
leveraging the knowledge transfer from pre-trained paraphrase estimation models~\cite{reimers:2019}.

In this work, we improve the \gls*{zsar} performance by employing two global semantic descriptors (\textit{i.e.,} descriptors computed over the whole video).
The first is based on object-action relationships, while the second is based on sentence-actions relationships.

\begin{figure}
\centerline{\includegraphics[width=0.98\linewidth]{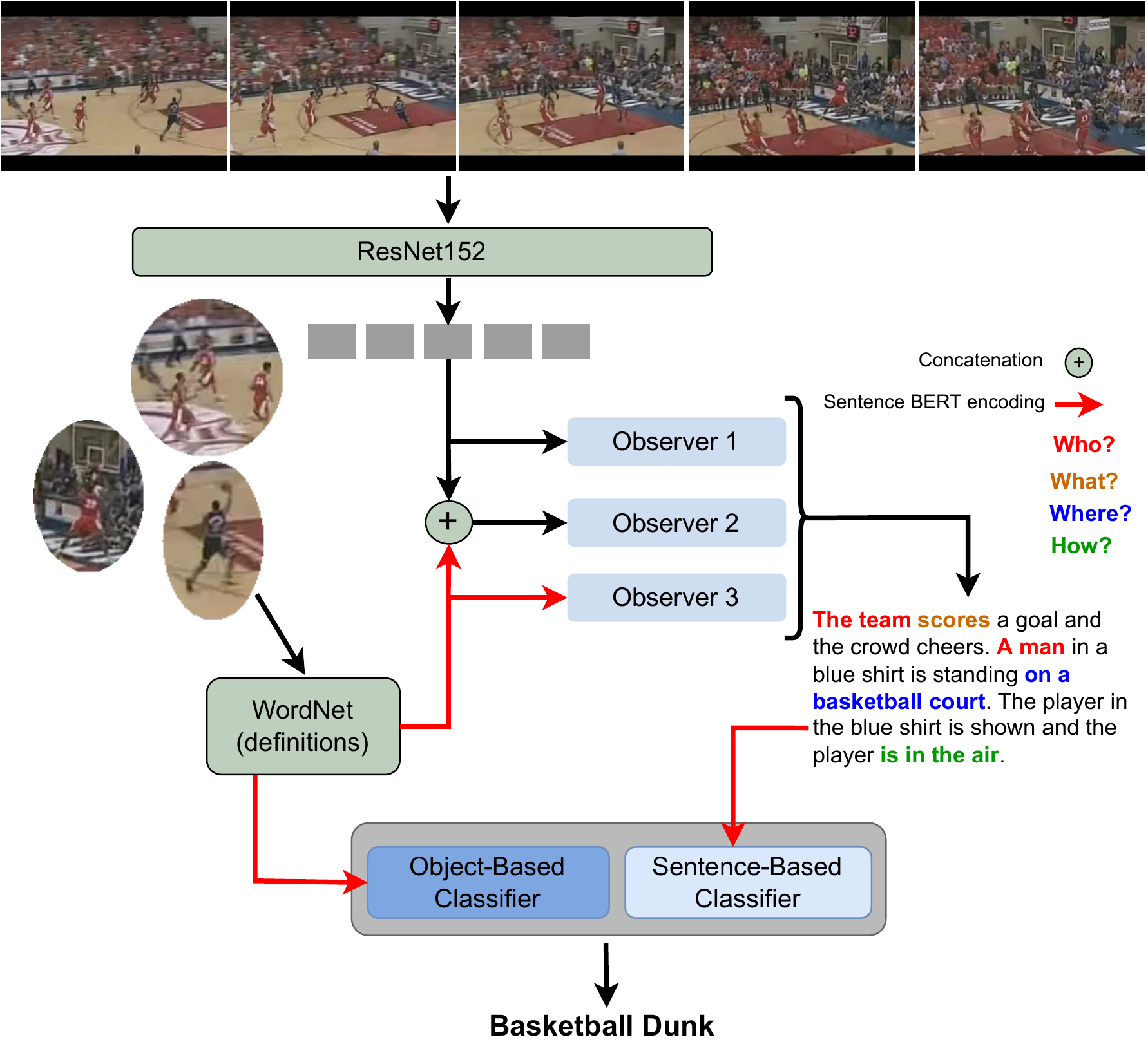}}

\vspace{-3mm}

\caption{Overview of the proposed method. 
We show the top-3 objects recognized in the video (left) and the WordNet component responsible for providing sentence definitions. We also show which features are fed to the observer models (\textit{i.e.,} the video captioning models), and the corresponding produced sentences~(right).}
\label{fig:model}
\end{figure}

\cref{fig:model}~(left) illustrates our object-based classifier, which uses a WordNet~\cite{fellbaum:1998} encoding to provide object definitions with natural language sentences.
\cref{fig:model}~(right) shows our sentence-based classifier, a network employed to classify a set of actions using a set of soft labeled sentences (\textit{i.e.}, annotated with a minimum human effort). In the \gls{zsar} inference step, this classifier is fed with sentences produced by video captioning methods, highlighted in \cref{fig:model} as \textit{Observers 1, 2 and~3}.

In summary, our main contributions are:
\textbf{(i)}~we demonstrate that object definitions and paraphrase embedding can improve \gls*{zsar} models based on object-affinity.
Our similarity matrices have fewer ambiguities than other methods;
\textbf{(ii)}~we demonstrate how textual descriptions can be used to learn a supervised action classifier based exclusively on semantic side information without hard human labeling (\textit{i.e.,} without labeling the sentences one by one).
Hence, we can generate sentences for each video (e.g., using video captioning~\cite{estevam:2021tell}) and feed this model to predict the corresponding action class.
In practice, these captioning models provide information on humans, objects, scenes, and their relationships, avoiding the need for manual definitions for affinity/prior functions on interactions while improving the performance; and, lastly, \textbf{(iii)}~the predictions using objects and sentences are easily combined to reach~\gls*{sota} performance on the Kinetics-400 dataset and competitive results on~UCF-101.
\section{Proposed Method}
\label{sec:proposedmethod}

Usually, the \gls*{zsar} goal is to classify samples belonging to a set of unseen action categories $ Z_{u} = {z_{1},...,z_{u_{n}}} $ (\textit{i.e.,} never seen before by the model) given a set of seen categories $ Z_{s} = {z_{1},...,z_{s_{n}}} $ as the training set. The problem is named ZSAR only if $Z_{u} \cap Z_{s} = \emptyset$.

Our work is even more restrictive because we do not use a seen set $ \mathcal{Z}_{s}$ with actions labeled for training our model;
this configuration has become popular in recent years~\cite{jain:2015,mettes:2017,mettes:2021,bretti:2021,estevam:2021tell}. Therefore, our goal is to classify unknown classes $Z_{u}$ using two types of semantic information on the videos: a textual description $s$ and a set of objects $Y$. They are independent of action labels, and the \gls*{zsar} restriction is respected. Our classifier for a video $v$ is given by

\begin{equation}
\label{eq:classifierdefinition}
\mathcal{C}(v) = \argmax_{z \in Z_{u}} (p_{sz} + \sum_{y \in Y}{p_{vy}g_{yz}}) \, ,
\end{equation}
\noindent where $p_{sz} \forall z \in Z_{u}$ is the classification score of a textual description over the set of unseen classes $Z_{u}$ given by a supervised model, as described in Section~\ref{subsec:supervisedsentencemodel};
$p_{vy} \forall y \in Y$ are the classification scores of objects given by an off-the-shelf classifier pre-trained in the ImageNet dataset; finally, $g_{yz}$~$\forall y \in Y$~and~$\forall z \in Z_{u}$
is an affinity score, that is, a term computed to estimate which objects are most related to which actions, inspired by Jain \textit{et al.}~\cite{jain:2015}, but with significant improvements as described in detail in Section~\ref{subsec:objectmodel}.

\subsection{Sentence-based Classifier}
\label{subsec:supervisedsentencemodel}

Unlike previous works~\cite{mishra:2018, mishra:2020}, where synthesized features were used for training supervised models, we project a classifier based exclusively on the semantic side information. Our classifier requires a set of descriptive sentences labeled with the corresponding action class label. We adopt the sentences from Estevam \textit{et al.}~\cite{estevam:2021tell} because they collected textual descriptions from the Internet and processed them to select a set of sentences closely related to each class name. This procedure proved beneficial for classification using the nearest neighbor rule due to the sentence embedder employed~\cite{reimers:2019}, and can be used to soft labeling individual sentences. 
Therefore, using the sentences from~\cite{estevam:2021tell}, we create a dataset $\mathcal{D} = \{S,Z_{u}\}$ with  sentence embedding-action label pairs and compute the probability $p_{sz}$ as
\begin{equation}
\label{eq:sentenceclassifier}
p_{sz} = \textit{softmax}(\text{GeLU}(sW + b)) \, ,
\end{equation}
\noindent where $s$ is the sentence embedding given by the \gls{sbert} model outputs~\cite{reimers:2019}, $\textit{softmax}$ returns a probability estimation on the $Z_{u}$ classes, $\text{GeLU}$ is a usual Gaussian Error Linear Unit, $W$ is an internal weight matrix, and $b$ is a bias~vector.

\subsection{Object-based Classifier}
\label{subsec:objectmodel}

First, we encode a video $v$ by the classification scores to the $m$ = $|Y|$ object classes from the object recognition model~\cite{beyer:2021} trained on ImageNet~\cite{krizhevsky:2017}.
\begin{equation}
\label{eq:objectencoding}
\textit{\textbf{p}}_{v} = [p(y_1|v),...,p(y_m|v)] \, ,
\end{equation}
\noindent where $p(y|v)$ is computed by averaging the logits over a set of video frames at 1~FPS. Then, we estimate the probabilities with a softmax layer.

We employ a common strategy to compute the affinity between an object class $y$ and action class $z$, enabling us to identify the most meaningful objects to describe an action.
Then, a translation of actions 
$z \in Z_{u}$ in terms of objects $y \in Y$ is given by
\begin{equation}
\label{eq:objectaffinity}
g_{yz} = s(w(y))^{T}s(z) \, ,
\end{equation}
\noindent or, in other terms, $\mathbf{g}_z = [s(w(y_1))...(s(w(y_m)))]^{T}s(z)$. In our case, $w(\cdot)$ returns the WordNet definition for the object label, and $s(\cdot)$ returns the 
Sentence-BERT~\cite{reimers:2019} encoding. This encoding does not require the Fisher vector computation on the individual words and, combined with object and sentence descriptions, conduct us to a higher performance than other object-based methods, as our results show.

\subsection{Sparsity}
\label{subsec:sparsity}

We sparsify $p_{sz}$, $p_{yz}$ and $g_{z}$ due to the performance improvements demonstrated in~\cite{jain:2015}. Formally, we redefine the original array as
\begin{equation}
\label{eq:sparsityobjpred}
\boldsymbol{\mathit{\hat{p}_{v_y}}} = [p(y_1,v)\delta(y_1,T_{v_y}),...,p(y_m,v)\delta(y_m,T_{v_y})]
\end{equation}
\begin{equation}
\label{eq:sparsitysentpred}
\boldsymbol{\mathit{\hat{p}_{s_{z}}}} = [p(z_1,v)\delta(z_1,T_{v_z}),...,p(z_n,v)\delta(z_n,T_{v_z})]
\end{equation}
\begin{equation}
\label{eq:sparsityaffinity}
\boldsymbol{\mathit{\hat{g}_z}} = [g_{zy_1}\delta(y_1, T_z),...,g_{zy_m}\delta(y_m,T_z)] \, ,
\end{equation}
\noindent where $\delta(.,T_{v_y})$, $\delta(.,T_{v_z})$ and $\delta(.,T_{z_y})$ are indicator functions, returning, 1 if class $y$ is among the top $T_{v_y}$ object classes in Equation~\ref{eq:sparsityobjpred}; returning 1 if class $z$ is among the top $T_{v_z}$ action classes in Equation~\ref{eq:sparsitysentpred}, and returning 1 if object class $y$ is in $T_{z_{y}}$~classes in Equation~\ref{eq:sparsityaffinity}, and 0 otherwise. $T_{v_y}$, $T_{v_z}$, and $T_{z_y}$ are parameters.
\section{Datasets and Evaluation Protocol}

Our experiments were conducted on the UCF-101~\cite{soomro:2012} and Kinetics-400~\cite{carreira:2017} datasets.
UCF101 is composed of $13{,}320$ videos from $101$ action classes, sampled at $25$ \gls{fps} and with an average duration of $7.2$s.
On the other hand, Kinetics-400 comprises~$306{,}245$~videos from $400$ action classes with at least 400 clips per class, collected from YouTube.
Each clip has a duration of $10$s.
As the videos came from YouTube, we were able to download only $242{,}658$ clips (\textit{i.e.,} $\approx$ 80\%) of the original dataset.
The videos have various frame rates and~resolutions. 

We encode the videos using two types of semantic information: objects and sentences.
For object encoding, we use the ResNet152 model from the Big Transfer (BiT) project~\cite{beyer:2021} pre-trained on ImageNet considering $21{,}843$ object classes.
For sentence encoding, we retrained the Transformer-based observers~\cite{estevam2021dense} from Estevam \textit{et al.}~\cite{estevam:2021tell} on the ActivityNet Captions dataset~\cite{krishna:2017}, without any class label from ActivityNet, replacing their i3D features with our ResNet152 features.
These features are sampled at each second after standardizing the videos to $25$~\gls{fps}.

We evaluate our model using accuracy and following two protocols for the UCF-101 dataset: conventional and TruZe~\cite{gowda:2021b}. The conventional protocol consists of splitting the dataset into seen and unseen classes. However, as explained in Section~\ref{sec:proposedmethod}, we do not use any class from the seen set, and the evaluated configurations are $0$\%/$50$\%, $0$\%/$20$\%, and $0$/$100$\%.
This protocol enables a fair comparison with other methods that use objects such as~\cite{jain:2015,mettes:2017,mettes:2021,bretti:2021,mettes:2022}. Due to being more restrictive, we consider that the comparison of our method with conventional methods such as~\cite{mandal:2019,gao:2019,kim:2021,chen:2021,zhu:2018,brattoli:2020,kerrigan:2021} is fair. Hence, we highlight the number of training classes each model uses in each configuration.

Additionally, we evaluate our model under the TruZe protocol to provide a fair comparison with Estevam~\textit{et al.}~\cite{estevam:2021tell}, which is the only method using sentence descriptions generated with video captioning techniques in the \gls{zsar} literature. In the TruZe protocol, overlapping classes between UCF-101 and Kinetics-400 are removed, enabling comparisons with methods that use 3DCNNs pre-trained on~Kinetics-400.

Finally, we evaluate the performance on the Kinetics-400 dataset. We adopt the same configurations from~\cite{mettes:2021} (\textit{i.e.,} $0$/$25$, $0$/$100$ and $0$/$400$ classes).
When a random subset of classes is used, we perform the evaluations with $50$ runs in all the protocols and datasets and report the average results.
\section{Experiments and Discussion}

As shown in Table~\ref{tab:ucf101results}, our complete method presented a higher performance in the UCF-101 dataset than other approaches in the literature under three split configurations. Our results are impressive compared to highly sophisticated object-based methods that explore intra-frame information such as scenes, actors, and interactions using manual defined affinity/relationship functions~\cite{bretti:2021,mettes:2021}.
Even our object-based classifier evaluated separately showed competitive results against $51$/$50$ and $664$/$50$ approaches.
These results demonstrate the effectiveness of our approach and the need to include more semantic information in \gls{zsar}~methods.

\begin{table}[!htb]
\centering
\caption{Results on the UCF-101 dataset under different numbers of test classes.}
\vspace{-2mm}
\resizebox{0.975\linewidth}{!}{
\begin{tabular}{lcccc}
\toprule
\multirow{2}{*}{Model}                                       & \multicolumn{4}{c}{UCF-101 - Testing classes}                                                                 \\
                                                             & Train         &  101                     &  50                       & 20                        \\
\midrule
Jain~\textit{et al.}~\cite{jain:2015}~\tiny{(ICCV)}          & $-$      &  30.3                    &  $-$                      & $-$                       \\
Mettes and Snoek~\cite{mettes:2017}~\tiny{(ICCV)}            & $-$      &  32.8                    & 40.4~$\pm$~1.0            & 51.2~$\pm$~5.0            \\
Mettes~\textit{et al.}~\cite{mettes:2021}~\tiny{(IJCV)}      & $-$      &  36.3                    & 47.3                      & 61.1                      \\
Bretti and Mettes~\cite{bretti:2021}~\tiny{(BMVC)}           & $-$      &  39.3                    & 45.4~$\pm$~3.6            & $-$                       \\
Mishra~\textit{et al.}~\cite{mishra:2018}~\tiny{(WACV)}      & 51       &  $-$                     & 22.7~$\pm$~1.2            & $-$                       \\
Mishra~\textit{et al.}~\cite{mishra:2020}~\tiny{(Neurocomputing)} & 51       &  $-$                & 23.9~$\pm$~3.0            & $-$                       \\
Mandal~\textit{et al.}~\cite{mandal:2019}~\tiny{(CVPR)}      & 51       &  $-$                     & 38.3~$\pm$~3.0            & $-$                       \\
Gao~\textit{et al.}~\cite{gao:2019}~\tiny{(AAAI)}            & 51       &  $-$                     & 41.6~$\pm$~3.7            & $-$                       \\
Kim~\textit{et al.}~\cite{kim:2021}~\tiny{(AAAI)}            & 51       &  $-$                     & 48.9~$\pm$~5.8            & $-$                       \\
Chen and Huang~\cite{chen:2021}~\tiny{(ICCV)}                & 51       &  $-$                     & 51.8~$\pm~$2.9            & $-$                       \\
Zhu~\textit{et al.}~\cite{zhu:2018}~\tiny{(CVPR)}            & 200      &  34.2                    & 42.5~$\pm$~0.9            & $-$                       \\
Brattoli~\textit{et al.}~\cite{brattoli:2020}~\tiny{(CVPR)}  & 664      &  39.8                    & 48                        & $-$                       \\
Kerrigan~\textit{et al.}~\cite{kerrigan:2021}~\tiny{(NeurIPS)} & 664    &  40.1                    & 49.2                      & $-$                       \\
\midrule
Ours (objects)                                               & $-$      &  39.8                    & 49.4~$\pm$~4.0            & 60.0~$\pm$~8.5            \\
Ours (sentences)                                             & $-$      &  30.8                    & 41.1~$\pm$~3.3            & 53.4~$\pm$~6.7            \\ 
Ours (objects + sentences)                                   & $-$      &  \textbf{40.9}           & \textbf{53.1~$\pm$~3.9}   & \textbf{63.7~$\pm$~8.3}   \\ 
\bottomrule
\end{tabular}
}
\label{tab:ucf101results}
\end{table}

Table~\ref{tab:ucf101truze} shows the results obtained in the UCF-101 datasets under the TruZe protocol.
To enable a fair comparison, we show the results from~Estevam~\textit{et al.}~\cite{estevam:2021tell} and include their pre-computed sentences in our model.
As expected, our sentence-based classifier, using sentences generated with ResNet152, produced results with lower accuracy than the version using sentences generated with i3D.
Surprisingly, this difference is only $2.7\%$ ($42.7$\% against $40.1$\%).
When compared to~\cite{estevam:2021tell}, the difference to our ResNet152 version is remarkable.
However, the  complete model achieves considerably better results. 
\begin{table}[!htb]
\centering
\caption{Results on the UCF-101 dataset under the TruZe protocol (34 classes for testing). Top-2 results are highlighted.}
\vspace{-2mm}
\begin{tabular}{lcc}
\toprule
\multirow{2}{*}{Model}                                                       & \multicolumn{2}{c}{UCF-101}                           \\
                                                                             & Train    & Accuracy (\%)             \\
\midrule
Wang and Chen~\cite{wang:2017} reported by~\cite{gowda:2021a}                & 67       & 16.0                      \\
Mandal~\textit{et al.}~\cite{mandal:2019} reported by~\cite{gowda:2021a}     & 67       & 23.4                      \\
Brattoli~\textit{et al.}~\cite{brattoli:2020} reported by~\cite{gowda:2021a} & 664      & 45.2                      \\
Gowda~\textit{et al.}~\cite{gowda:2021a}                                     & 67       & 45.3                      \\
Estevam~\textit{et al.}~\cite{estevam:2021tell}                                  & $-$      & 49.1                      \\
\midrule
Ours (objects)                                                               & $-$      & 55.3                      \\
Ours (sentences as in~\cite{estevam:2021tell})                                   & $-$      & 42.7                      \\
Ours (objects + sentences~as in~\cite{estevam:2021tell})                         & $-$      & \textbf{60.5}             \\
\midrule
Ours (objects)                                                               & $-$      & 55.3                      \\
Ours (sentences)                                                             & $-$      & 40.1                      \\
Ours (objects + sentences)                                                   & $-$      & \textbf{57.0}             \\
\bottomrule
\end{tabular}
\label{tab:ucf101truze}
\end{table}

\begin{figure*}[!htb]
    \centering
    \captionsetup[subfigure]{captionskip=-0.25pt,font={scriptsize},justification=centering} 
    \subfloat[][]{
    \resizebox{0.5\linewidth}{!}{
    \includegraphics[height=10ex]{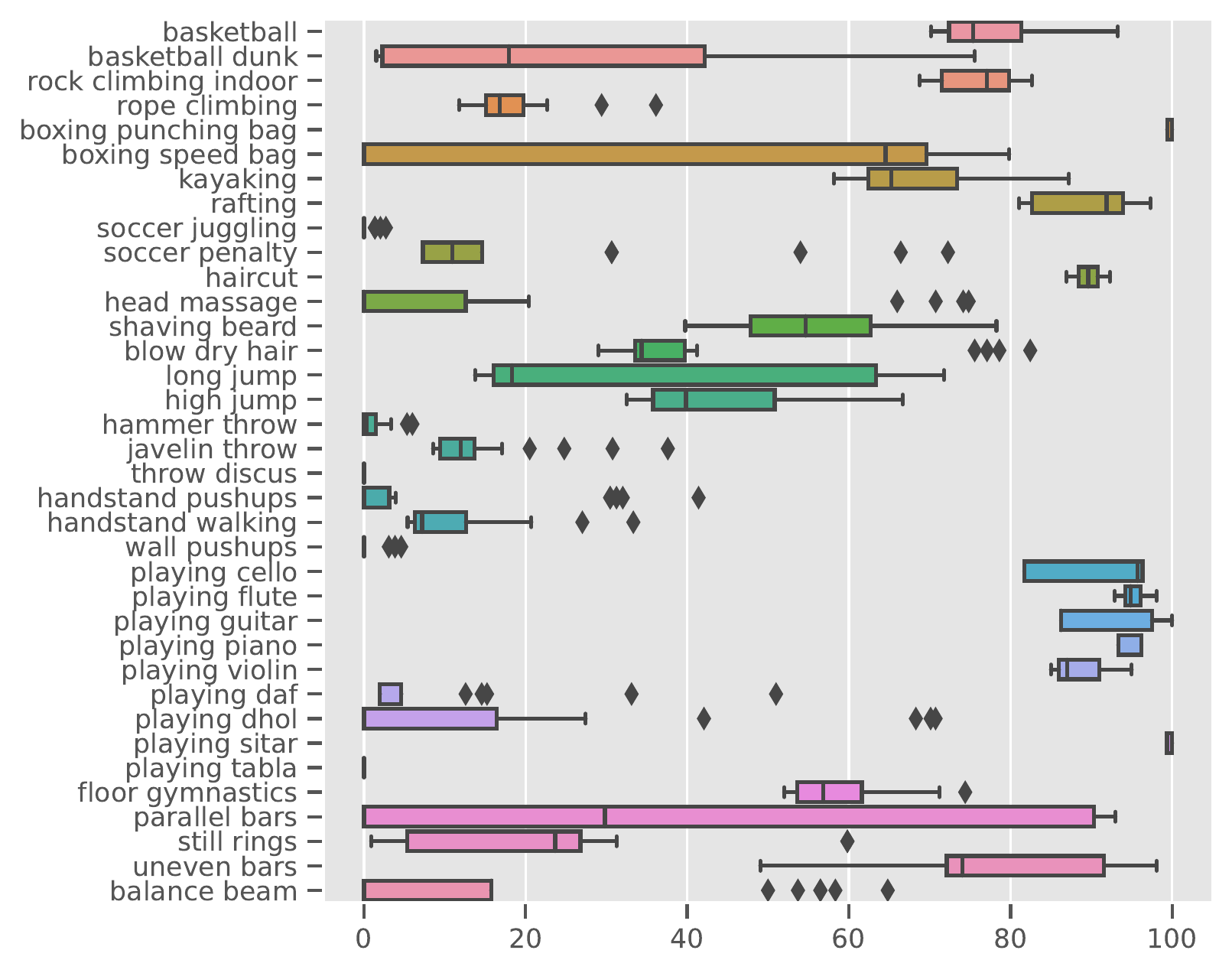}
    }}
    \subfloat[][]{
    \resizebox{0.5\linewidth}{!}{
    \includegraphics[height=10ex]{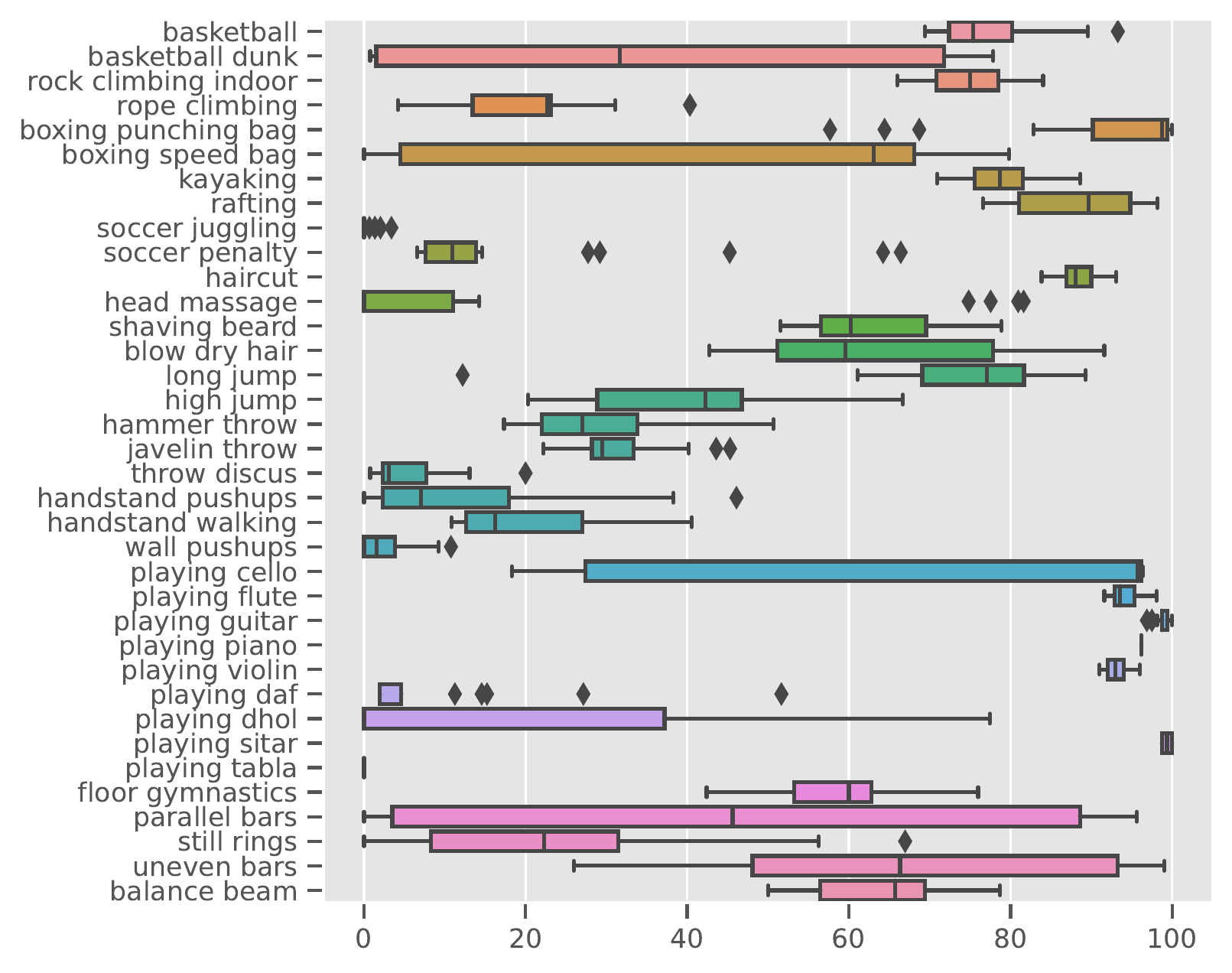}
    }
    }
    \vspace{-0.25mm}
    \caption{\textit{Per-class accuracy} computed over $50$ random runs on the UCF101 dataset for a subset of similar semantic classes. In (a) the results are shown for the object-based model and in (b) for the complete model.}
    \label{fig:boxplot}
\end{figure*}

The Kinetics-400 dataset is very challenging for~\gls{zsar}. There are several classes semantically similar to each other (\textit{e.g.,} eating [burger, cake, carrots, chips, doughnuts, hotdog, ice cream, spaghetti, watermelon] and juggling [balls, fire, soccer ball]).
Moreover, as several methods are trained with features pre-computed in this dataset, there is not a sufficiently large list of methods with which they can be compared.
In Table~\ref{tab:kineticsresults}, we present our results compared to~\cite{mettes:2021}, \cite{bretti:2021}, and \cite{mettes:2017}, which are object-based.
As can be observed, the inclusion of semantic information in the form of natural language embedded with \gls{sbert} improves the accuracy by around $40$\% to $50$\% in all configurations.
Surprisingly, the $0$/$400$ performance for the complete model was lower than that of the object-based classifier, contrary to the results obtained in all the other~experiments. We believe this occurred because the sentences produced with video captioning techniques were not sufficiently discriminative for similar actions.

\begin{table}[!htb]
\centering
\caption{Results on the Kinetics-400 dataset under different numbers of test classes. No classes were used for training. The best results are highlighted.}
\vspace{-2mm}
\resizebox{0.95\linewidth}{!}{
\begin{tabular}{lccc}
\toprule
\multirow{2}{*}{Model}                                                     & \multicolumn{3}{c}{Kinetics-400 - Testing classes}                                              \\
                                                                           & 400                       & 100               & 25                            \\
\midrule
Mettes and Snoek~\cite{mettes:2017}~\tiny{(ICCV)}                          & 6.0                       & 10.8~$\pm$~1.0    & 21.8~$\pm$~3.5                \\
Mettes~\textit{et al.}~\cite{mettes:2021}~\cite{mettes:2021}~\tiny{(IJCV)} & 6.4                       & 11.1~$\pm$~0.8    & 21.9~$\pm$~3.8                \\ 
Bretti and Mettes~\cite{bretti:2021}~\tiny{(BMVC)}                         & 9.8                       & 18.0~$\pm$~1.1    & 29.7~$\pm$~5.0                \\
\midrule
Ours (objects)                                                             & \textbf{20.4}             & 32.4~$\pm$~2.4    & 49.3~$\pm$~6.8                \\
Ours (sentences)                                                           & 13.3                      & 25.1~$\pm$~2.2    & 44.2~$\pm$~5.5                \\
Ours (objects + sentences)                                                 & 19.4                      & \textbf{35.1~$\pm$~2.4} & \textbf{54.6~$\pm$~6.1} \\
\bottomrule
\end{tabular}
}
\label{tab:kineticsresults}
\end{table}

\cref{fig:boxplot} illustrates a similar effect in the UCF101 dataset. We compute the \textit{per-class accuracy} for each action in each random run. Then, we produce the boxplot shown in the figure by grouping semantic similar classes. For instance, considering the classes ``\textit{basketball}'' and ``\textit{basketball dunk}'', they are not necessarily unknown in all runs. We observe that ``\textit{basketball dunk}'' varies from $0$ in some cases to around $70\%$ in others.
At the same time, ``\textit{basketball}'' shows lower variation in their per-class accuracy.
Hence, we conclude that the model is prone to predict ``\textit{basketball}'' when both classes are unknown.
The same behavior occurs between ``\textit{boxing punching bag}'' and ``\textit{boxing speed bag}'', and, also in other cases, as shown in the figure.
For some classes (\textit{e.g.}, ``\textit{handstand pushups}'', ``\textit{handstand walking}'', and ``\textit{playing dhol}''), we observe an increase in the performance shown by the increase in the bar length and a shift of the median. At the same time, ``\textit{playing cello}'' presents the worst performance.
\section{Conclusions}

In this work, we introduced a new~\gls*{zsar} model based on two global semantic descriptors. We demonstrated the effectiveness of adopting semantic information with sentences in natural language for both descriptors. Our supervised sentence classifier is considerably more straightforward than other supervised approaches in the literature (\textit{e.g.,}~\cite{mishra:2018, mishra:2020}) and presents a higher performance compared to them.
Additionally, our object-based classifier also benefits from sentences, thus reaching remarkable results compared to other object-based methods. In future work, we intend to investigate different semantic descriptors with a focus on improving semantically similar~classes, a problem that we still observe in our~method.

\balance

\ifalt
\bibliographystyle{IEEEtran-2}
\else
\bibliographystyle{IEEEtran}
\fi
\bibliography{refs}
\end{document}